\documentclass[11pt,a4paper]{article}
\usepackage[hyperref]{naaclhlt2019}
\usepackage{times}
\usepackage{latexsym}
\usepackage{url}
\usepackage{algorithm}
\usepackage{algpseudocode}
\usepackage{amssymb}
\usepackage{xcolor}

\newcommand{\predicate}[1]{{\fontfamily{cmtt}\selectfont \small{`#1'}}}

\usepackage[colorinlistoftodos]{todonotes}

\usepackage{textcomp}
\newcommand{\ftextnumero}{{\fontfamily{txr}\selectfont \textnumero}}

\aclfinalcopy %
\def\aclpaperid{7} %

\date{}

\title{Learning from Implicit Information in Natural Language Instructions\\ for Robotic Manipulations}

\author{
Ozan Arkan Can\Thanks{Equal contribution} \Thanks{ocan13@ku.edu.tr}  \\ Ko\c{c} University \\ \And
Pedro Zuidberg Dos Martires$^*$\Thanks{pedro.zuidbergdosmartires@cs.kuleuven.be} \\ KU Leuven \And
Andreas Persson \\ \"Orebro University  \\ \AND
Julian Gaal \\ Osnabr\"uck University \\ \And
Amy Loutfi \\ \"Orebro University \And
Luc De Raedt  \\ KU Leuven \AND
Deniz Yuret \\ Ko\c{c} University \And
Alessandro Saffiotti \\ \"Orebro University
}
\date{September 2018}

\usepackage{natbib}
\usepackage{graphicx}
\usepackage{amsmath}
\usepackage{nicefrac}
\usepackage{microtype}

\newenvironment{talign}
 {\align}
 {\endalign}

\begin{document}
\frenchspacing
\maketitle

\begin{abstract}
Human-robot interaction often occurs in the form of instructions given from a human to a robot. For a robot to successfully follow instructions, a common representation of the world and objects in it should be shared between humans and the robot so that the instructions can be grounded. Achieving this representation can be done via learning, where both the world representation and the language grounding are learned simultaneously. However, in robotics this can be a difficult task  due to the cost and scarcity of data. In this paper, we tackle the problem by separately learning the world representation of the robot and the language grounding. While this approach can address the challenges in getting sufficient data, it may give rise to inconsistencies between both learned components. %
Therefore, we further propose Bayesian learning to resolve such inconsistencies between the natural language grounding and a robot's  world representation by exploiting spatio-relational information that is implicitly present in instructions given by a human. Moreover, we demonstrate the feasibility of our approach on a scenario involving a robotic arm in the physical world.  
\end{abstract}

\section{Introduction}\label{sec:intro}
Consider yourself standing in your kitchen and having your robot assist you in preparing tonight's meal. You then give it the instruction: \textit{`fetch the bowl next to the bread knife!'}. For the robot to correctly perform your intended instruction, which is grounded in your world representation, it must correctly ground your natural language instruction into its own world representation.

This small scenario already introduces the two key components of language grounding in robotics: the construction of a world representation from sensor data and the grounding of natural language into the constructed representation. Ideally these two components would be learned in a joint fashion \cite{hu2017learning, johnson2017inferring, santoro2017simple, hudson2018compositional, perez2018film}. However, the scarcity of data makes this approach impractical. The millions of data points necessary for state-of-the-art joint computer vision and natural language processing are simply non-existing. We opt, therefore, to separately learn the world representation component and the language grounding component.

One approach for constructing a world representation of a robot is through so-called \textit{perceptual anchoring}. Perceptual anchoring handles the problem of creating and maintaining, over time, the correspondence between symbols in a constructed world model and perceptual data that refer to the same physical object~\cite{coradeschi&saffiotti-2000}. In this work, we use sensor driven bottom-up anchoring~\cite{loutfi.et.al-2005}, whereby anchors (symbolic representations of objects) can be created by perceptual observations derived directly from the input sensory data. When modeling a scene, based on visual sensor data, through object anchoring, noise and uncertainties will inevitably be present. This leads, for example, to a  green 'apple' object being incorrectly anchored as a 'pear'.

For the language grounding, we opt to perform the learning on \textbf{synthetic data} that simulates the world represented as anchors.
This means that we do not ground the language using sensor data as signal but a symbolic representation of the world. During training these symbols are synthetic and simulated, and during the deployment of the language grounding these are anchors provided by an anchoring system. As the real world is inherently relational and as natural language instructions are often given in terms of spatial relations as well, the learned language grounder must also be able to ground spatial language such as \textit{`next to'}.

As a result of learning the construction of a world model and the language grounding separately, \textbf{contradictions} arise \textbf{between} the world \textbf{representations} of a human and a robot. The supervision that an instruction would give to a robot is not present when learning the representation of the world of a robot. These inconsistencies then propagate through to inconsistencies between the instructions a human gives to a robot and the robot's world model. To ensure that a robot is able to correctly carry out an instruction, such inconsistencies must be resolved and the world model of the robot be matched to the world model of the human.

This is not the first paper that tackles the problem of belief revision in robotics. However, prior work \cite{tellexll2013toward,thomason2015learning,she2017interactive}, with the notable exception of~\cite{mast2016probabilistic}, relied on explicit information transfer between humans and robots when inconsistencies arose in grounded language and the robot's world representation. An example would be a robot asking  clarification questions until it is clear what the human meant \cite{tellexll2013toward}. 

We propose an approach that probabilistically reasons over the grounding of an instruction and a robot's world representation in order to perform Bayesian learning to update the world representation given the grounding. This is closely related to the work of \citeauthor{mast2016probabilistic} who also deploy a Bayesian learning approach. The key difference, however, is that they do not learn the language component but ground a description of a scene by relying on a predefined model to ground language.
We demonstrate the validity of our approach for reconciling instructions and world representations on a showcase scenario involving a camera, a robot arm and a natural language interface.

\section{Preliminaries}
\label{sec:preliminaries}

The overarching objective of our system is to plan and execute robot manipulation actions based on natural language instructions. Presumptuously, this requires, in the first place, that both the planner of the robot manipulator, as well as the natural language grounder (cf. Section~\ref{sec:nlg}), share a joint semantically rich object-centered model of the perceived environment, i.e., a \textit{semantic world model}~\cite{elfring.et.al-2013}. 

\subsection{Visual Object Anchoring}
\label{sec:anchoring}

In order to model a semantic object-centered representation of the external environment, we rely upon the notions and definitions found within the concept of perceptual anchoring \cite{coradeschi&saffiotti-2000}. Following the approach for sensor-driven bottom-up acquisition of perceptual data, as described by \cite{persson2019semantic}, the used anchoring procedure is, initially, triggered by sensory input data provided by a \textit{Kinect2 RGB-D sensor}. Each frame of input \textit{RGB-D} data is, subsequently, processed by a \textit{perceptual system}, which exploits both the visual $2{\text -}D$ information, as well as the $3{\text -}D$ depth information, in order to: \textit{1)} detect and segment the subset of data (referred to as \textit{percepts}), that originates from a single individual object in the physical world, and \textit{2)} measure \textit{attribute values} for each segmented percept, e.g., measuring a \textit{position attribute} as the $\mathbb{R}^3$ geometrical center of an object, or a visual \textit{color attribute} measured as a color histogram (in \textit{HSV} color space). 

The percept-symbol correspondence is, thereafter, established by a \textit{symbolic system}, which handles the grounding of measured attributes values to corresponding predicate symbols through the use of \textit{predicate grounding relations}, e.g., a certain peek in a color histogram, measured as a \textit{color attribute}, is mapped to a corresponding predicate symbol \predicate{red}. In addition, we promote the use of an \textit{object classification} procedure in order to semantically categorize and label each perceived object. The convolutional neural network (CNN) architecture that we use for this purpose is based on the \textit{GoogLeNet} model \cite{szegedy.et.al-2015}, which we have trained and fine-tuned based on $101$ object categories that can be expected to be found in a kitchen domain.

The extracted perceptual and symbolic information for each perceived object is then encapsulated in an internal data structure $\alpha^x_t$, called an \emph{anchor}, indexed by time $t$ and identified by a unique identifier $x$ (e.g. \predicate{mug-2}, \predicate{apple-4}, etc.). The goal of an \textit{anchoring system} is to manage these anchors based on the result of a \textit{matching function} that compares the attribute values of an unknown candidate object against the attribute values of all previously maintained anchors. Anchors are then either created or maintained through two general functionalities:

\begin{itemize}

    \item \textit{Acquire} -- initiates a new anchor whenever a candidate object is received that does not match any existing anchor $\alpha^x$. 

    \item \textit{Re-acquire} -- extends the definition of a matching anchor $\alpha^x$ from time $t-k$ to time $t$. This functionality assures that the percepts pointed to by the anchor are the most recent perceptual (and consequently also symbolic) representation of the object.
 
\end{itemize}

However, comparing attribute values of anchored objects and percepts by some distance measure and deciding, based on the measure, whether an unknown object has previously been perceived or not is a non-trivial task.
Nevertheless, since anchors are created or maintained through either one of the two principal functionalities \textit{acquire} and \textit{re-acquire}, it is evident that the desired outcome for the combined compared values is a \textit{binary} output, i.e. should a percept be acquired or re-acquired. In previous work on anchoring \cite{persson2019semantic}, we have therefore suggested that the problem of invoking a correct anchoring functionality is a problem that can be approximated through learning from examples and the use of \textit{classification algorithms}. For this work, we follow the same approach. 

\subsection{Natural Language Grounding}\label{sec:nlg}
\begin{figure}[ht!]
	\begin{center}
	\resizebox{\columnwidth}{0.47\textheight}{
	    \includegraphics{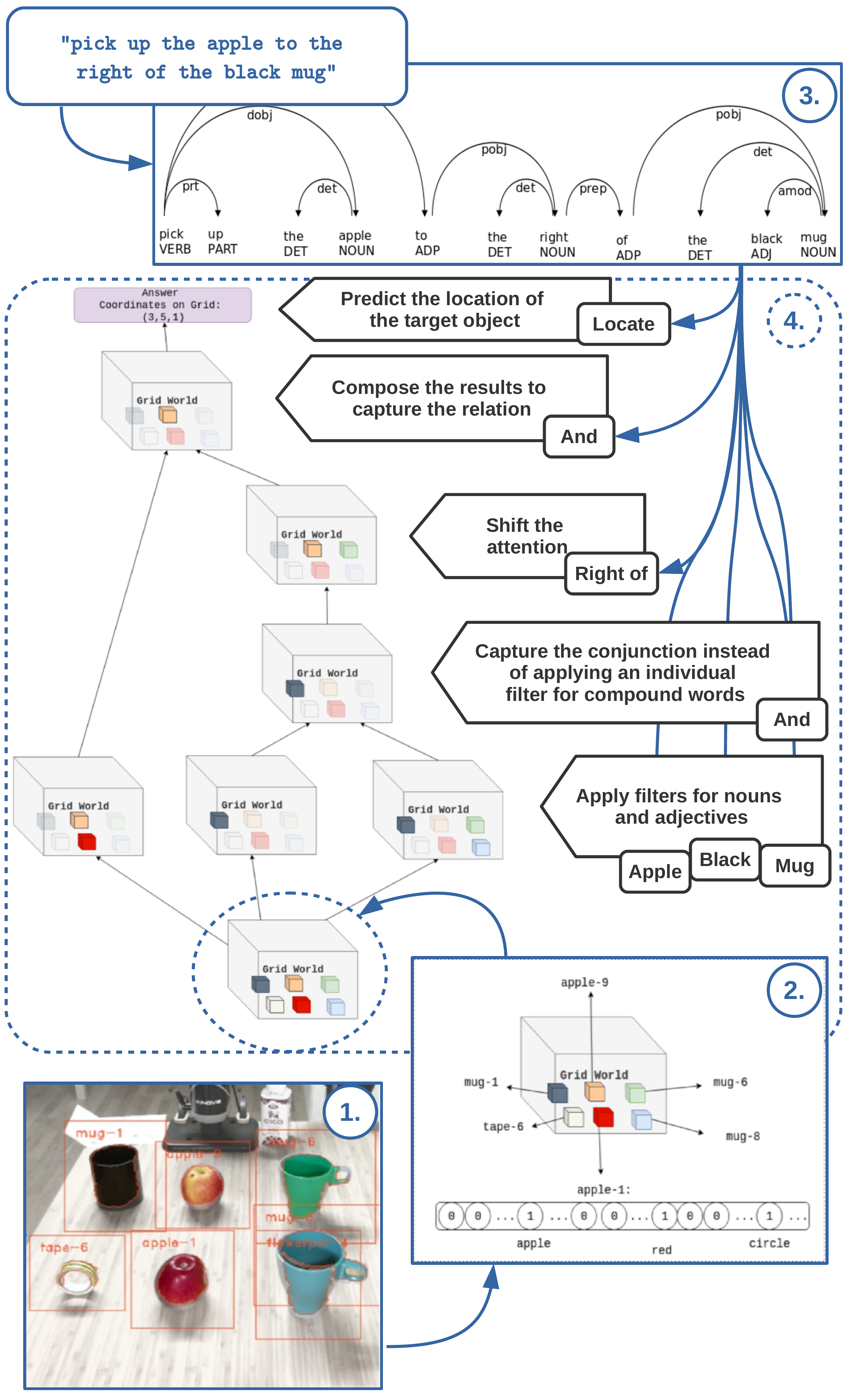}
	    }
	\end{center}
    \caption{Demonstration of the language grounding process for the instruction "pick up the apple to the right of the black mug". Anchoring system sends the snapshot of the anchors (1). Then, a preprocessor transforms the anchors into a grid representation which the language grounding system operates on (2). The parser parses the given instruction and generates a computation graph which specifies the execution order of neural modules (3). Finally, the neural modules are executed according to the computation graph to produce the action (4).}
    \label{lgmain}
\end{figure}

In this study, we focus on understanding spatial language that includes \emph{pick up} and \emph{place} related verbs, and \emph{referring expressions}. An instruction refers to a target object using its representative features (e.g. color, shape, size). If a noun phrase does not resolve the ambiguity in the world, the instruction resolves the ambiguity by specifying the target object with its relative position to other surrounding objects. This hierarchy tries to bring the attention to finding the unique object, then shifts the attention to the targeted object. Based on this idea, we model the language grounding process as controlling the attention on the world representation by adapting the neural module networks approach proposed by \newcite{andreas2016neural}.

Our natural language grounder has three components: a preprocessor, an instruction parser and a program executor. Given specific anchor information (Figure~\ref{lgmain} -- \ftextnumero~1), the preprocessor transforms the anchor information into an intermediate representation in grid form (Figure~\ref{lgmain} -- \ftextnumero~2). The instruction parser produces a computational program  by exploiting the syntactic representation (Figure~\ref{lgmain} -- \ftextnumero~3) of the instruction with a dependency parser\footnote{https://spacy.io/}. The program executor runs (Figure~\ref{lgmain} -- \ftextnumero~4) the program on the intermediate representation to produce commands.

\textbf{Preprocessor.} The anchoring framework maintains the object descriptions predicted from the raw visual input. To be able to ground the language onto those descriptions, we map the available information (object class, color, size and shape attributes) to a 4D grid representation. We represent each anchor as a multi-hot vector and assign this vector to a cell where the real world coordinates of the object fall into.

\textbf{Program Executor.} The program generated by the parser is a collection of neural components that are linked to each other depending on the computation graph. The design of neural components reflects our intuition about the attention control. A \textbf{\emph{Detect}} module is a convolutional neural network with a learnable filter that captures a noun or an adjective. This module creates an attention map over the input grid.
\begin{equation}
Detect(w,b,x) = relu(w \otimes x + b)
\end{equation}
The \emph{Detect} module operates on the original grid input $x$ tensor, where the dimensions are $(W, H, L, C)$. The first three dimensions represent the spatial dimensions and $C$ denotes the length of the feature vector. $w$ is the filter of size $(1,1,1,C)$ and $b$ is the bias. $\otimes$ is a convolution operation.

Although a \emph{Detect} module can capture the meaning of a noun phrase (e.g., red book), the model cannot generalize to unseen compound words. To overcome this, we design the \emph{And} module to compose the output of incoming modules. This module multiplies the inputs element-wise in order to calculate the composition of words (e.g., the big red book). Since the incoming inputs are attention maps over the grid world, an \textbf{\emph{And}} module produces a new attention map by taking the conjunction of its inputs. In the following equation, the $\odot$ denotes the element-wise multiplication.
\begin{equation}
And(a_1, a_2) = a_1 \odot a_2    
\end{equation}
An output of a subgraph for a noun phrase is an attention map that highlights the positions for the corresponding objects that occur. A \textbf{\emph{Shift}} module shifts this attention in the direction of the preposition that the module represents. This module is also a convolutional neural network similar to a \textbf{\emph{Detect}} module. However, the module remaps the attention instead of capturing the patterns in the grid world.
\begin{equation}
Shift(w,a) = relu(w \otimes a)
\end{equation}
The \emph{Shift} module operates on an incoming attention map, where the dimensions are $(W, H, L, 1)$. $w$ is the filter of size of $(2*W+1, 2*H+1, 2*L+1, 1)$. We use the padding to be able to perform the shifting operation over the whole grid. The pad size is the same as the input size.

A \textbf{\emph{Locate}} module takes an attention map and produces a probability distribution over cells by applying a softmax classifier for being the targeted object. We use the cell with the highest probability as the prediction. A \textbf{\emph{Position}} module gets a source anchor, a preposition and a target anchor, and produces a real world coordinate. It merely calculates the position available in the direction of the preposition from the target anchor, where the source anchor can fit.  

\textbf{Parser.} We find the verbs in the instruction along with the subtrees attached to them. For each verb and its subtree, we search for the direct object of the verb. Then we build a subgraph for the direct object and its modifiers. Depending on the verb type, we build different subgraphs. If the verb is "pick up" related, then we look for the preposition that relates the given noun to another noun. If one is found, then a subgraph is created for the preposition object using the noun phrase that the object belongs to. Finally, the end point of the subgraph is combined with a \emph{Shift} module. For each preposition object, we repeat the same process to handle prepositional phrase chains.

\begin{figure}[ht!]
	\begin{center}
	\resizebox{\columnwidth}{0.45\textheight}{
		\includegraphics{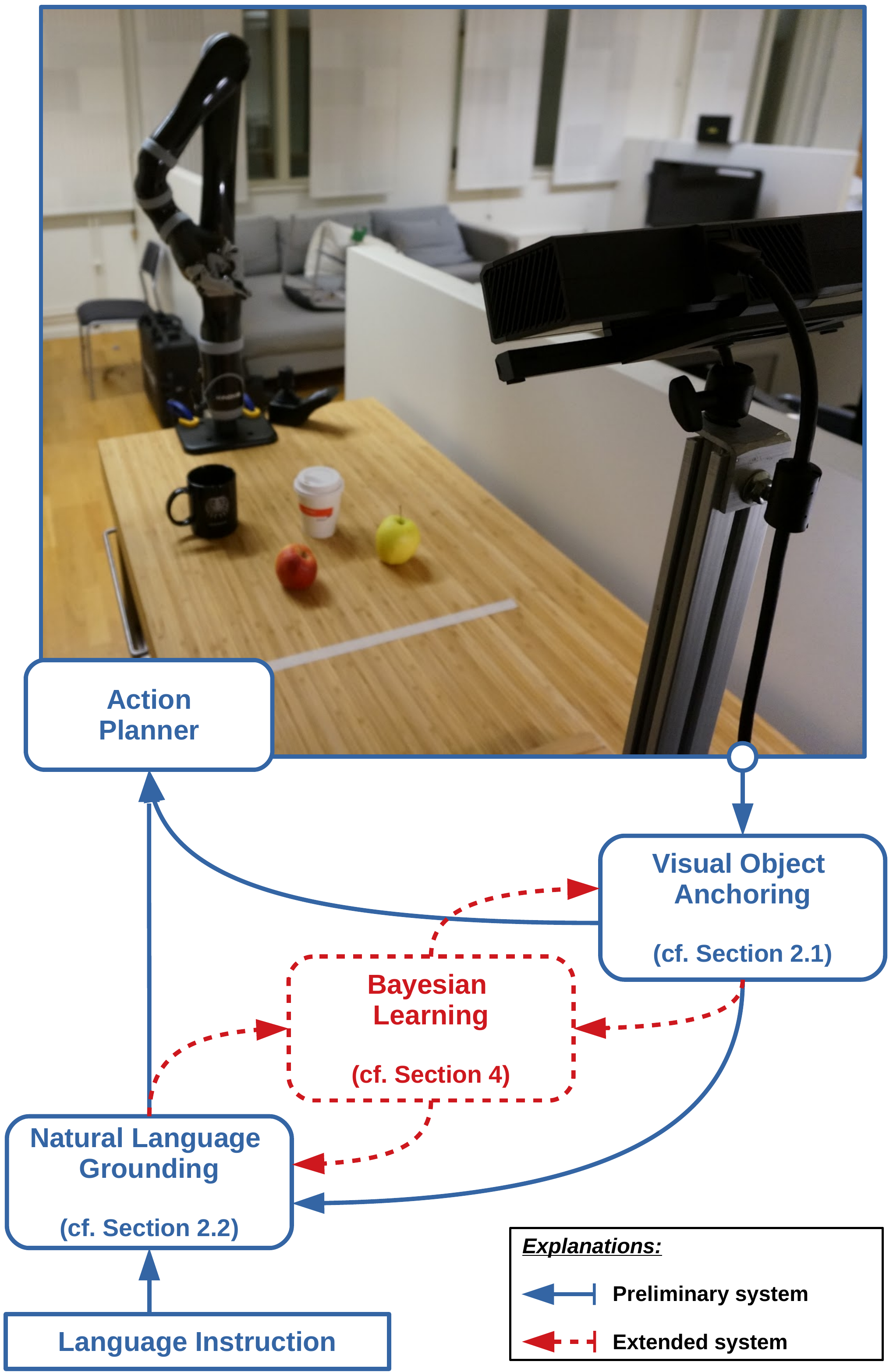}
		}
	\end{center}
    \caption{A depiction of both used physical system setup (upper), as well as used software architecture (lower). The arrows represent the flow of data between the modules of the software architecture. Blue solid arrows and boxes illustrate the preliminary system (outlined in Section~\ref{sec:preliminaries}), while red dashed arrows and boxes illustrate the novel extension for reasoning about different symbolic label configurations (and hence resolving inconsistencies between language and perception), by using Bayesian learning (as presented in Section~\ref{sec:resolving}).}
    \label{fig:reground_setup}
\end{figure}

If the verb is "put" related, we find the preposition that is linked to the verb and the object of the preposition. We build a subgraph that refers to the object of the preposition similar to the "pick up" case. Finally, there is a \emph{Position} module to produce the coordinates to put the direct object, where the position is referred with the auxiliary objects.

\section{System Description}
\label{sec:system}

In the upper part of Figure~\ref{fig:reground_setup}, we illustrate our physical \textit{kitchen table} system setup, which consists of the following devices: \textit{1)} a \textit{Kinova Jaco light-weight manipulator}~\cite{campeau2019kinova}, \textit{2)} a \textit{Microsoft Kinect2 RGB-D} sensors, and \textit{3)} a dedicated PC with an Intel\textsuperscript{\textcopyright} Core\textsuperscript{\texttrademark} i7-6700 processor and an Nvidia GeForce GTX 970 graphics card.

In addition, we have a modularized software architecture that utilizes the libraries and communication protocols available in the Robot Operating System (ROS)\footnote{http://www.ros.org/}. Hence, each of the modules, illustrated in the lower part of Figure~\ref{fig:reground_setup}, consists of one or several individual subsystems (or ROS nodes). For example, the \textit{visual object anchoring} module consists of the following subsystems: 1) a \textit{perceptual system}, 2) a \textit{symbolic system}, and 3) an \textit{anchoring system}. For a seamless integration between software and hardware, we are further taking advantage of both the \textit{MoveIt! Motion Planning Framework}\footnote{https://moveit.ros.org/}, as well as the \textit{ROS-Kinect2 bridge} developed by \cite{iai_kinect2}. The \textit{MoveIt! "planning scene"} of the \textit{action planner} for the robot manipulator, as well as the grid world representation used by the language grounding system (cf. Section~\ref{sec:nlg}), are, subsequently, both populated by the same updating anchoring representations (cf. Section~\ref{sec:anchoring}). Hence, the visual sensory input stream is indirectly mapped to both objects considered in the dialogue by the language grounder, as well as the objects upon which actions are executed.

\section{Resolving Inconsistencies} \label{sec:resolving}
Based purely on the perceptual input, the anchoring system produces a probability distribution $p(l)$ over the possible labels (e.g. $[0.65:\text{apple},0.35:\text{pear}]$) for each anchor. We are now interested in the probability of a label $l$ for an anchor given a natural language instruction $i$ and the grounding $g$ of that instruction in the real world. This is the conditional probability $p(l \mid g,i)$. We introduce, furthermore, the notion of a label configuration $c$. This is easiest explained by an example: imagine having two anchors and each of the anchors has two possible labels, then there are $2\times 2$ possible label configurations. A label configuration is, hence, a label assignment to all the anchors present in the scene.

Now we need to transform the conditional probability into a function that is computable by the anchoring system and the language grounder. The first steps (Equations \ref{al1:1}-\ref{al1:3}) are quite straight forward and follow basic probability calculus. 
\begin{talign}\textstyle{}
    p(l|g,i)&=\sum_c p(l,c|g,i) \label{al1:1}\\
    &=\sum_c p(l|c,g,i)p(c|g,i)\\
    &=\sum_c p(l|c)p(c|g,i) \label{al1:3}
\end{talign}
In Equation \ref{al1:3} we assume that $g$ and $i$ are conditionally independent of the label of an anchor given the label configuration $c$. This can be seen in the following way. Imagine two anchors with two possible labels each. Given that we are in a specific label configuration, we immediately know what label the single anchors have. This means that the probability of a label for an anchor is $1$ if it matches the label in the configuration and $0$ otherwise. This reasoning is independent of the grounding and the instruction.  

We have now split up the labels (produced by the anchoring system) and the grounding into two factors, which can be calculated separately. The first one can be calculated as follows:
\begin{align}
    p(l\mid c) = \frac{p(l,c)}{p(c)} = \frac{\prod_{j\in c} p(l_j)}{N_c} \label{al1bis:1}
\end{align}
This is the product of the probabilities of the labels that constitute a label configuration divided by the number of configurations. Assuming a uniform distribution over the label configurations (division by $N_c$) is equivalent to assuming that each possible label configuration is equally likely \textit{a priori}. This means that we make no assumption about which class of objects occur more regularly or which class of objects (of the 101 possible classes) occur more often together with other classes of objects.

We tackle now the second factor in Equation \ref{al1:3}. Equation \ref{al2:1}-\ref{al2:4} are again straightforward probability calculus. In Equation \ref{al2:5} we assume that the label configuration and the instruction are independent: their probabilities factorize. In Equation \ref{al2:6} the probabilities of $i$ cancel out and we assume again a uniform distribution for the label configurations (cf Equation \ref{al1bis:1}). In Equation \ref{al2:7} we then have a numerator and denominator that are expressed in terms of $p(g\mid c, i)$, which is exactly the function approximated by our neural language grounding system, cf. subsection \ref{sec:nlg}.
\begin{align}\textstyle
    p(c|g,i)&=\frac{p(c,g,i)}{p(g,i)}\label{al2:1}\\
    &=\frac{p(g|c,i)p(c,i)}{p(g,i)}\\
    &=\frac{p(g|c,i)p(c,i)}{\sum_c p(c,g,i)}\\
    &=\frac{p(g|c,i)p(c,i)}{\sum_c p(g|c,i)p(c,i)} \label{al2:4}\\
    &=\frac{p(g|c,i)p(c)p(i)}{\sum_c p(g|c,i)p(c)p(i)}\label{al2:5}\\
    &=\frac{p(g|c,i)\nicefrac{1}{N_c}}{\sum_c p(g|c,i)\nicefrac{1}{N_c}}\label{al2:6}\\
    &=\frac{p(g|c,i)}{\sum_c p(g|c,i)}\label{al2:7}
\end{align}

Plugging Equations \ref{al1bis:1} and \ref{al2:7} back into Equation \ref{al1:3} gives the learned probability of the label $l$ of an anchor given the instruction $i$ and the grounding $g$ of that instruction.
\begin{talign}
    p(l \mid g,i ) = \frac{\sum_c \left(\prod_{j\in c}p(l_j) \right) p(g|c,i)}{N_c \sum_c p(g|c,i)}\label{eqn:label_learn}
\end{talign}

As mentioned in Section \ref{sec:anchoring} the anchoring system encapsulates $101$ object categories, which means that the anchoring system produces a categorical probability distribution over $101$ different labels for each anchor. With only two anchors this results in already $101^2$ different configurations. It is easy to see that computing $p(l \mid g,i)$ (cf. Equation \ref{eqn:label_learn}) suffers from this curse of dimensionality. Therefore, we limited ourselves to the two labels with the highest probability per anchor, in the experiments too. This gives $2^{N_A}$ possible configurations, with $N_A$ being the number of anchors present.   

\section{Experiments}

\subsection{Synthetic Data} \label{sec:synthetic}
Data demanding nature of neural networks requires large amounts of data to generalize well. Artificial data generation is one way of generating such datasets  \cite{andreas2016neural, kuhnle2017shapeworld, johnson2016clevr}. Therefore, we designed a series of artificial learning tasks before applying the model to a real-world problem. In each task, we generate a random grid world that provides the necessary complexity and ambiguity that fit the scenario. First, an object is placed on the grid world and decorated with attributes randomly as the target  object. Then depending on the scenario, an auxiliary object and distractors (objects that have similar attributes as the target object) are placed on the grid world. We also generate objects that are not related to the target (or auxiliary object) to introduce additional noise. We limit the total number of objects to $10$. We set the number of distractors as $2$ in the experiments. Finally, we generate the ground truth computation graph for composing neural modules. We list the scenarios below in increasing order of difficulty (i.e., a combination of the ambiguity present in the grid world and the number of language components involved). 

\begin{enumerate}
    \item Using the {\bf name} of a targeted object in the instruction is enough to localize the targeted object.
    \item There is more than one object that has the same category with a targeted object. To solve the ambiguity, one or more discriminative {\bf adjective(s)} are used.
    \item The same world configuration as the second one. To solve the ambiguity, the object is described with a {\bf prepositional phrase} that utilizes a single referent object.
    \item The same world configuration as the third. {\bf Adjectives} are used to describe a targeted object in addition to a {\bf prepositional phrase}. In this case, adjectives are unnecessary, but the scenario measures whether additional components bring noise or not. 
    \item All other objects that have the same category with a targeted object have the same set of features as the targeted object has. Hence, the targeted object is only distinguishable by its position. To solve the ambiguity, the object is described with a \textbf{prepositional phrase} that utilizes a referent object along with necessary \textbf{adjectives}.
    \item It is a {\bf random} scenario from the above list.
\end{enumerate}

\subsection{Training}
\begin{figure}[ht!]
	\begin{center}
	\resizebox{\columnwidth}{0.22\textheight}{
		\includegraphics{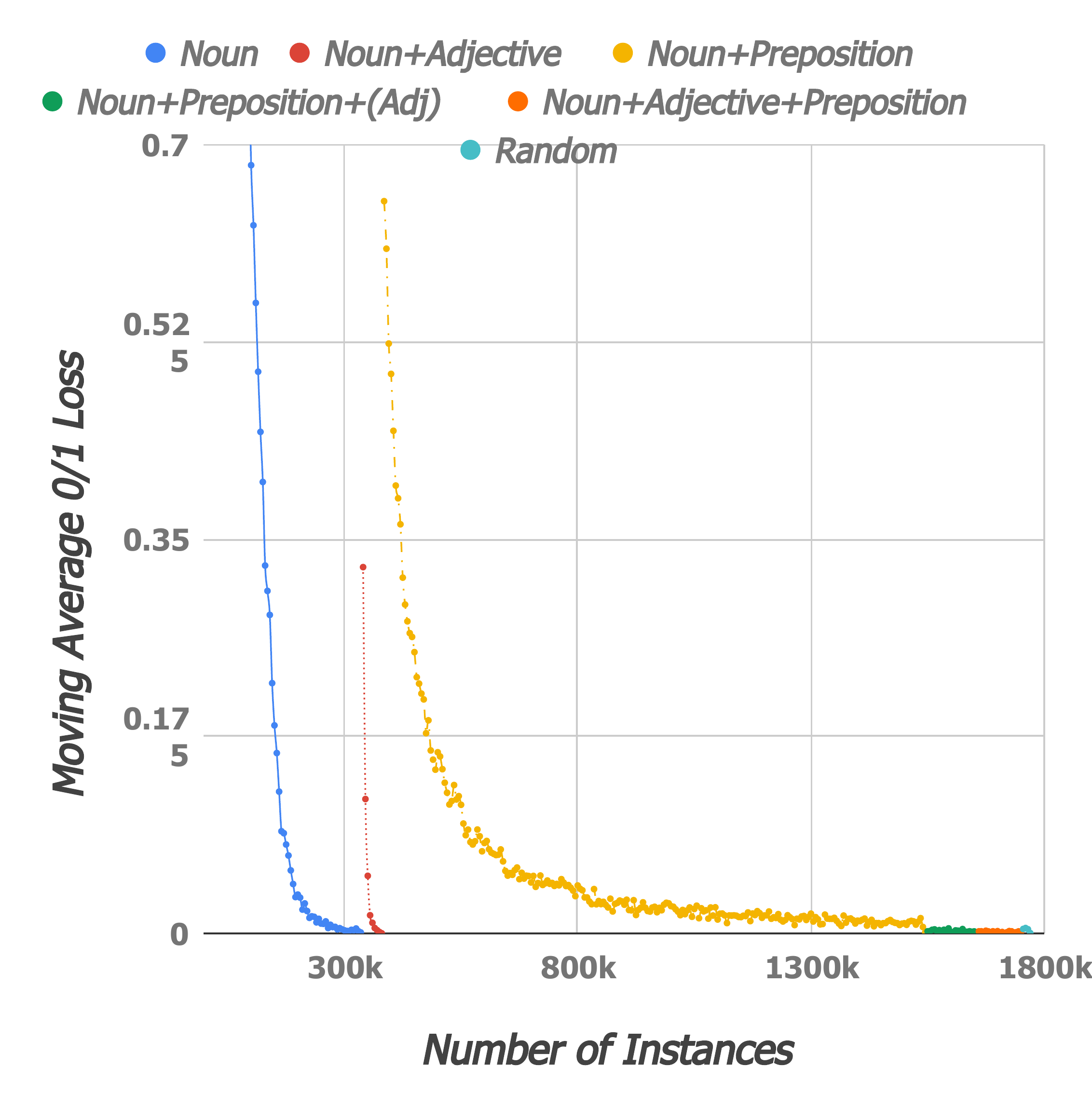}
	}
	\end{center}
    \caption{Learning curve of the neural modules.}
    \label{fig:learningcurve}
\end{figure}
To be able to measure the compositionality of learned modules, we have two different settings for the data generation. For training, we constrain the 75\% of possible attributes for an object class and locations on the grid world that an instance of that object class can present. During testing, we use unconstrained samples generated for the same scenario. This way, we can evaluate if the model infers unseen word compositions, e.g. inferring red mug after seeing red book and black mug in the training time. We follow a curriculum schema to train our modules. Starting from the first scenario described in Section \ref{sec:synthetic}, we train the model on a stream of constrained randomly generated samples. We evaluate the model periodically on unconstrained samples generated for each period and continue training until the moving average error on the test data falls under a threshold (e.g. 1e-5 in our experiments). We then continue to train the model for the next scenario using learned weights. We set the number of nouns, adjectives and prepositions as 102, 26, and 27, respectively to match with the anchoring system. We use Adam \cite{adam} with default parameters (i.e. $lr=0.001$, $\beta_1=0.9$, $\beta_2=0.999$) for the optimization.  

Figure \ref{fig:learningcurve} presents the learning curve of the model. The third graph (yellow) demonstrates that learning prepositions requires more data as compared to learning nouns (first graph) or adjectives (second graph). The reason for this behavior is twofold. First, the \emph{Shift} modules have more weights to be learned than the \emph{Detect} modules. Second, while the \emph{Detect} modules have a one to one mapping between input and output, the \emph{Shift} modules have many to many relations. There might be more than one active area in the input of a \emph{Shift} module. Since it needs to remap highlighted areas on the grid to other areas
, it needs to see different examples that occur in different parts of the grid world in order to learn to ignore the position of the active area.

\begin{figure*}[ht!]
	\begin{center}
		\includegraphics[width=0.9\textwidth]{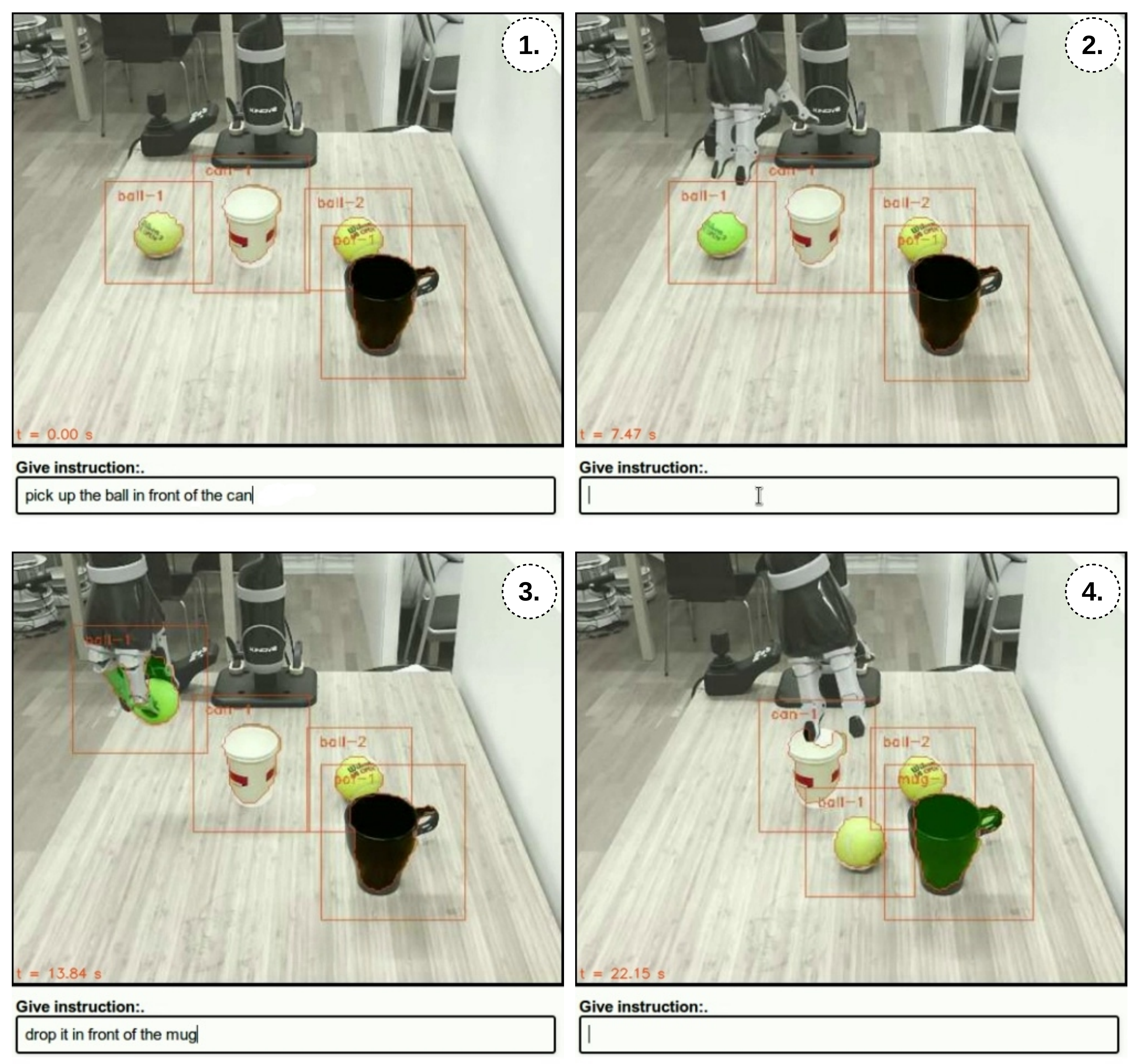}

	\end{center}
    \caption{We give the robot the instruction: \textit{"pick up the ball in front of the can"}. The robot executes the action and waits for further instructions. We then give the instruction to \textit{"drop it in front of the mug"}. The problem in this step is that there is no object classified as \predicate{mug}, which means that none of the objects has as label with highest probability \predicate{mug}. We correct for this through probabilistic reasoning over not only the top label for each object but a number of top ranked labels per object. This allows the anchoring system to correct its classification of an object based on what we as humans think an object is. Given the instruction, the anchoring system re-classifies the black object from \predicate{pot} to \predicate{mug}. The instruction is then successfully carried out. The recorded video can be found here: \url{https://vimeo.com/302072685}.}
    \label{fig:showcase}
\end{figure*}

The remaining graphs show the effectiveness of our design to compose learned modules. Since we do not train any modules from scratch, we can handle the composition of nouns, adjectives and prepositions effectively. Since it is the first time we train all components together in scenario 4, the training requires more data than one would expect when compared with graphs 5 (orange) and 6 (cyan).

\section{Showcase}

We now proceed with a demonstration of the integrated system: we have a Kinect camera that observes the world, the anchoring representation that builds up a representation of the world based on the raw image data, the language grounder that takes as input a natural language instruction and a probabilistic reasoning component that resolves possible inconsistencies between the robot's world representation and the instruction.

The physical setup up is identical to the one depicted in the image  in Figure \ref{fig:reground_setup}: the robot arm is mounted on the opposite site of a kitchen table of the Kinect camera. The natural language instruction is passed to the language grounder via an \textit{instruction prompt}. In each of the four panels in Figure \ref{fig:showcase}, the instruction prompt is seen at the bottom as rectangular box. We further describe the scenario in the caption of Figure \ref{fig:showcase}.

\section{Related Work}

Our work is related to two research domains: modular neural nets for language grounding and human-robot interaction for handling ambiguities in one or more modalities. \citet{andreas2016neural,andreas2016learning} introduced neural module networks for visual question answering. \citet{johnson2017inferring, hu2017learning} developed policy gradient based approaches to learn to generate layouts instead of using a dependency parser based method. \citet{hu2017modeling, yu2018mattnet, cirik2018using} applied modular neural networks approach on `Referring Expression Understanding` task. \citet{das2018neural} demonstrated the usage of neural module networks in decision taking in a simulated environment. To our knowledge, the present study is the first work that uses neural module networks approach in the real-world robotic setting.

Learning from human interaction has been extensively studied. \citet{lemaignan2011you, lemaignan2012grounding} developed a cognitive architecture that makes decisions by using symbolic information provided as facts (pre-defined) or extended via human-robot dialogues. When compared with our system, their system neither operates on the sensory input nor deals with the uncertainty in the world. \citet{tellexll2013toward} proposed a system to ask questions to disambiguate the ambiguities presented in the instructions. The robot decides the most ambiguous part of the command which is defined based on a metric derived from  entropy and asks questions about it to reduce the uncertainty. They update the generalized grounding graph \cite{kollar2013generalized} with answers obtained from the user and use these to perform inference. In contrast, we fix the ambiguity present in the perceptual data. \citet{she2017interactive} proposed a system to learn to ask questions during the learning of verb semantics. They work on the \textit{Tell me Dave environment} \cite{misra2014tell}. The work represents the environment as grounded state fluents (i.e. a weighted logic representation). In this work, language grounding is modeled as the difference between before and after state for an action sequence. They modeled the interactive learning as an MDP and solved it with reinforcement learning.

\citet{thomason2015learning} proposed a system that learns the meaning of natural language commands through human-robot dialog. They represent the meaning of instructions with $\lambda$-calculus semantic representation. Their semantic parser starts with an initial knowledge and learns through training examples generated by the human-robot conversations. Their dialog manager is a static policy which generates questions from a discrete set of action, patient, recipient tuples. \citet{padmakumar2017integrated} improved this work with a learnable dialog manager. They train both the dialog manager and the semantic parser with reinforcement learning. This approach was further extended in \cite{thomason2019improving}, where the authors combine the approach in \citet{thomason2015learning} and \citet{thomason2017opportunistic} to obtain a system that is capable of concept acquisition through clarification dialogues. Instead of asking questions, we implicitly fix the perception with the information hidden in instructions. A further difference to these works is that we learn the language component in a simulated offline step, whereas they deploy active online learning, starting from a limited initial vocabulary. 

This is also related to the work of \citeauthor{perera2015quantity}, who present a system that tries to emulate child language learning strategies by describing scenes to a robot agent, which has to learn actively new concepts. The authors deploy probabilistic reasoning to manage erroneous sensor readings in the vision system. Apart from the active learning approach, there is also a conceptual difference: in our work, we do not consider discrepencies between the perceptual system (anchoring) and the language grounder as errors in the perceptual system but simply as different models of the world.\footnote{This view taps into the philosophical question of whether one can ever truly know the nature of an object, cf. \textit{thing-in-itself}~\cite{kant1878prolegomena}, for which we omit a discussion.}

As mentioned in Section \ref{sec:intro}, the work related closest to our approach is presented in \citet{mast2016probabilistic}. The authors base their work on geometric \textit{conceptual spaces}  \cite{gardenfors2004conceptual}, which situates their work in the sub-domain of \textit{top-down anchoring}~\cite{coradeschi&saffiotti-2000}. The geometric conceptual spaces induce a probabilistic model-based language grounder. This enables a robot to reason probabilistically over a description of a scene, given by an other agent, and single out the object that is most likely being referred to. In contrast, we present an approach to perform Bayesian learning over a learned language grounding model and a bottom-up anchoring approach. 

\section{Conclusions and Future Work}
We introduced the problem of belief revision in robotics based solely on implicit information available in natural language in the setting of sensor-driven bottom-up anchoring in combination with a learned language grounding model. This is in contrast to prior works, which study either explicit information or are based on top-down anchoring. We proposed a Bayesian learning approach to solve the problem and demonstrated its validity on a real world showcase involving computer vision, natural language grounding and robotic manipulation.

In future work we would like to perform a more quantitative analysis of our approach to which end it is imperative to circumvent the curse of dimensionality emerging in the Bayesian learning step (cf. Equation \ref{eqn:label_learn}). It would also be interesting to investigate whether our approach is amenable to natural language other than instructions. 

A main limitation of our current approach is the limited  size  of the predefined vocabulary. It would be more practicable if a robot were able to extend its vocabulary through the interaction with a human, i.e. through dialogue. A possible solution would be to learn a probabilistic model (which resolves inconsistencies between language and vision) that takes into account the possible of currently unknown vocabulary occurring. Such an approach would still allow us to learn the anchoring of objects and the language grounding separately, while learning a much richer model to resolve inconsistencies than the one described in this work.

\section*{Acknowledgements}
This work has been supported by the ReGROUND project (http://reground.cs.kuleuven.be), which is a  CHIST-ERA project funded by the EU H2020 framework program, the Research Foundation - Flanders, the Swedish Research Council (Vetenskapsr{\aa}det), and the Scientific and Technological Research Council of Turkey (TUBITAK). The work is also supported by Vetenskapsr{\aa}det under the grant number: 2016-05321 and by  TUBITAK under the grants 114E628 and 215E201.

\bibliographystyle{acl_natbib}
\bibliography{references}

\begin{thebibliography}{35}
\expandafter\ifx\csname natexlab\endcsname\relax\def\natexlab#1{#1}\fi

\bibitem[{Andreas et~al.(2016{\natexlab{a}})Andreas, Rohrbach, Darrell, and
  Klein}]{andreas2016learning}
Jacob Andreas, Marcus Rohrbach, Trevor Darrell, and Dan Klein.
  2016{\natexlab{a}}.
\newblock Learning to compose neural networks for question answering.
\newblock \emph{arXiv preprint arXiv:1601.01705}.

\bibitem[{Andreas et~al.(2016{\natexlab{b}})Andreas, Rohrbach, Darrell, and
  Klein}]{andreas2016neural}
Jacob Andreas, Marcus Rohrbach, Trevor Darrell, and Dan Klein.
  2016{\natexlab{b}}.
\newblock \href
  {https://www.cv-foundation.org/openaccess/content_cvpr_2016/html/Andreas_Neural_Module_Networks_CVPR_2016_paper.html}
  {Neural module networks}.
\newblock In \emph{Proceedings of the IEEE Conference on Computer Vision and
  Pattern Recognition}, pages 39--48.

\bibitem[{Campeau-Lecours et~al.(2019)Campeau-Lecours, Lamontagne, Latour,
  Fauteux, Maheu, Boucher, Deguire, and L'Ecuyer}]{campeau2019kinova}
Alexandre Campeau-Lecours, Hugo Lamontagne, Simon Latour, Philippe Fauteux,
  V{\'e}ronique Maheu, Fran{\c{c}}ois Boucher, Charles Deguire, and
  Louis-Joseph~Caron L'Ecuyer. 2019.
\newblock Kinova modular robot arms for service robotics applications.
\newblock In \emph{Rapid Automation: Concepts, Methodologies, Tools, and
  Applications}, pages 693--719. IGI Global.

\bibitem[{Cirik et~al.(2018)Cirik, Berg-Kirkpatrick, and
  Morency}]{cirik2018using}
Volkan Cirik, Taylor Berg-Kirkpatrick, and Louis-Philippe Morency. 2018.
\newblock Using syntax to ground referring expressions in natural images.
\newblock In \emph{Thirty-Second AAAI Conference on Artificial Intelligence}.

\bibitem[{Coradeschi and Saffiotti(2000)}]{coradeschi&saffiotti-2000}
S.~Coradeschi and A.~Saffiotti. 2000.
\newblock Anchoring symbols to sensor data: preliminary report.
\newblock In \emph{Proc.\ of the 17th {AAAI} Conf.}, pages 129--135, Menlo
  Park, CA. AAAI Press.
\newblock Online at http://www.aass.oru.se/\~{}asaffio/.

\bibitem[{Das et~al.(2018)Das, Gkioxari, Lee, Parikh, and
  Batra}]{das2018neural}
Abhishek Das, Georgia Gkioxari, Stefan Lee, Devi Parikh, and Dhruv Batra. 2018.
\newblock Neural modular control for embodied question answering.
\newblock \emph{arXiv preprint arXiv:1810.11181}.

\bibitem[{Elfring et~al.(2013)Elfring, van~den Dries, van~de Molengraft, and
  Steinbuch}]{elfring.et.al-2013}
J.~Elfring, S.~van~den Dries, M.J.G. van~de Molengraft, and M.~Steinbuch. 2013.
\newblock \href {https://doi.org/http://dx.doi.org/10.1016/j.robot.2012.11.005}
  {Semantic world modeling using probabilistic multiple hypothesis anchoring}.
\newblock \emph{Robotics and Autonomous Systems}, 61(2):95--105.

\bibitem[{G{\"a}rdenfors(2004)}]{gardenfors2004conceptual}
Peter G{\"a}rdenfors. 2004.
\newblock \emph{Conceptual spaces: The geometry of thought}.
\newblock MIT press.

\bibitem[{Hu et~al.(2017{\natexlab{a}})Hu, Andreas, Rohrbach, Darrell, and
  Saenko}]{hu2017learning}
Ronghang Hu, Jacob Andreas, Marcus Rohrbach, Trevor Darrell, and Kate Saenko.
  2017{\natexlab{a}}.
\newblock Learning to reason: End-to-end module networks for visual question
  answering.
\newblock In \emph{Proceedings of the IEEE International Conference on Computer
  Vision}, pages 804--813.

\bibitem[{Hu et~al.(2017{\natexlab{b}})Hu, Rohrbach, Andreas, Darrell, and
  Saenko}]{hu2017modeling}
Ronghang Hu, Marcus Rohrbach, Jacob Andreas, Trevor Darrell, and Kate Saenko.
  2017{\natexlab{b}}.
\newblock Modeling relationships in referential expressions with compositional
  modular networks.
\newblock In \emph{Proceedings of the IEEE Conference on Computer Vision and
  Pattern Recognition}, pages 1115--1124.

\bibitem[{Hudson and Manning(2018)}]{hudson2018compositional}
Drew~A Hudson and Christopher~D Manning. 2018.
\newblock Compositional attention networks for machine reasoning.
\newblock \emph{arXiv preprint arXiv:1803.03067}.

\bibitem[{Johnson et~al.(2016)Johnson, Hariharan, van~der Maaten, Fei-Fei,
  Zitnick, and Girshick}]{johnson2016clevr}
Justin Johnson, Bharath Hariharan, Laurens van~der Maaten, Li~Fei-Fei,
  C~Lawrence Zitnick, and Ross Girshick. 2016.
\newblock Clevr: A diagnostic dataset for compositional language and elementary
  visual reasoning.
\newblock \emph{arXiv preprint arXiv:1612.06890}.

\bibitem[{Johnson et~al.(2017)Johnson, Hariharan, van~der Maaten, Hoffman,
  Fei-Fei, Lawrence~Zitnick, and Girshick}]{johnson2017inferring}
Justin Johnson, Bharath Hariharan, Laurens van~der Maaten, Judy Hoffman,
  Li~Fei-Fei, C~Lawrence~Zitnick, and Ross Girshick. 2017.
\newblock Inferring and executing programs for visual reasoning.
\newblock In \emph{Proceedings of the IEEE International Conference on Computer
  Vision}, pages 2989--2998.

\bibitem[{Kant(1878)}]{kant1878prolegomena}
Immanuel Kant. 1878.
\newblock \emph{Prolegomena zu einer jeden K{\"u}nftigen Metaphysik: die als
  Wissenschaft wird auftreten konnen}.
\newblock Verlag von Leopold Voss.

\bibitem[{Kingma and Ba(2014)}]{adam}
Diederik Kingma and Jimmy Ba. 2014.
\newblock Adam: A method for stochastic optimization.
\newblock \emph{arXiv preprint arXiv:1412.6980}.

\bibitem[{Kollar et~al.(2013)Kollar, Tellex, Walter, Huang, Bachrach,
  Hemachandra, Brunskill, Banerjee, Roy, Teller et~al.}]{kollar2013generalized}
Thomas Kollar, Stefanie Tellex, Matthew~R Walter, Albert Huang, Abraham
  Bachrach, Sachi Hemachandra, Emma Brunskill, Ashis Banerjee, Deb Roy, Seth
  Teller, et~al. 2013.
\newblock \href
  {https://people.csail.mit.edu/sachih/home/wp-content/uploads/2014/04/G3_JAIR.pdf}
  {Generalized grounding graphs: A probabilistic framework for understanding
  grounded language}.
\newblock \emph{JAIR}.

\bibitem[{Kuhnle and Copestake(2017)}]{kuhnle2017shapeworld}
Alexander Kuhnle and Ann Copestake. 2017.
\newblock Shapeworld-a new test methodology for multimodal language
  understanding.
\newblock \emph{arXiv preprint arXiv:1704.04517}.

\bibitem[{Lemaignan et~al.(2011)Lemaignan, Ros, Alami, and
  Beetz}]{lemaignan2011you}
S{\'e}verin Lemaignan, Raquel Ros, Rachid Alami, and Michael Beetz. 2011.
\newblock What are you talking about? grounding dialogue in a perspective-aware
  robotic architecture.
\newblock In \emph{2011 RO-MAN}, pages 107--112. IEEE.

\bibitem[{Lemaignan et~al.(2012)Lemaignan, Ros, Sisbot, Alami, and
  Beetz}]{lemaignan2012grounding}
S{\'e}verin Lemaignan, Raquel Ros, E~Akin Sisbot, Rachid Alami, and Michael
  Beetz. 2012.
\newblock Grounding the interaction: Anchoring situated discourse in everyday
  human-robot interaction.
\newblock \emph{International Journal of Social Robotics}, 4(2):181--199.

\bibitem[{Loutfi et~al.(2005)Loutfi, Coradeschi, and
  Saffiotti}]{loutfi.et.al-2005}
A.~Loutfi, S.~Coradeschi, and A.~Saffiotti. 2005.
\newblock Maintaining coherent perceptual information using anchoring.
\newblock In \emph{Proc.\ of the 19th {IJCAI} Conf.}, pages 1477--1482,
  Edinburgh, UK.

\bibitem[{Mast et~al.(2016)Mast, Falomir, and Wolter}]{mast2016probabilistic}
Vivien Mast, Zoe Falomir, and Diedrich Wolter. 2016.
\newblock Probabilistic reference and grounding with pragr for dialogues with
  robots.
\newblock \emph{Journal of Experimental \& Theoretical Artificial
  Intelligence}, 28(5):889--911.

\bibitem[{Misra et~al.(2014)Misra, Sung, Lee, and Saxena}]{misra2014tell}
Dipendra~K Misra, Jaeyong Sung, Kevin Lee, and Ashutosh Saxena. 2014.
\newblock \href
  {http://citeseerx.ist.psu.edu/viewdoc/summary?doi=10.1.1.716.7045} {Tell me
  dave: Contextsensitive grounding of natural language to mobile manipulation
  instructions}.
\newblock In \emph{in RSS}. Citeseer.

\bibitem[{Padmakumar et~al.(2017)Padmakumar, Thomason, and
  Mooney}]{padmakumar2017integrated}
Aishwarya Padmakumar, Jesse Thomason, and Raymond~J Mooney. 2017.
\newblock \href {https://par.nsf.gov/biblio/10025851} {Integrated learning of
  dialog strategies and semantic parsing}.
\newblock In \emph{Proceedings of the 15th Conference of the European Chapter
  of the Association for Computational Linguistics (EACL)}.

\bibitem[{Perera and Allen(2015)}]{perera2015quantity}
Ian Perera and James Allen. 2015.
\newblock Quantity, contrast, and convention in cross-situated language
  comprehension.
\newblock In \emph{Proceedings of the Nineteenth Conference on Computational
  Natural Language Learning}, pages 226--236.

\bibitem[{Perez et~al.(2018)Perez, Strub, De~Vries, Dumoulin, and
  Courville}]{perez2018film}
Ethan Perez, Florian Strub, Harm De~Vries, Vincent Dumoulin, and Aaron
  Courville. 2018.
\newblock Film: Visual reasoning with a general conditioning layer.
\newblock In \emph{Thirty-Second AAAI Conference on Artificial Intelligence}.

\bibitem[{Persson et~al.(2019)Persson, Zuidberg Dos~Martires, Loutfi, and
  De~Raedt}]{persson2019semantic}
Andreas Persson, Pedro Zuidberg Dos~Martires, Amy Loutfi, and Luc De~Raedt.
  2019.
\newblock Semantic relational object tracking.
\newblock \emph{arXiv preprint arXiv:1902.09937}.

\bibitem[{Santoro et~al.(2017)Santoro, Raposo, Barrett, Malinowski, Pascanu,
  Battaglia, and Lillicrap}]{santoro2017simple}
Adam Santoro, David Raposo, David~G Barrett, Mateusz Malinowski, Razvan
  Pascanu, Peter Battaglia, and Timothy Lillicrap. 2017.
\newblock A simple neural network module for relational reasoning.
\newblock In \emph{Advances in neural information processing systems}, pages
  4967--4976.

\bibitem[{She and Chai(2017)}]{she2017interactive}
Lanbo She and Joyce Chai. 2017.
\newblock \href {http://www.aclweb.org/anthology/P17-1150} {Interactive
  learning of grounded verb semantics towards human-robot communication}.
\newblock In \emph{Proceedings of the 55th Annual Meeting of the Association
  for Computational Linguistics (Volume 1: Long Papers)}, volume~1, pages
  1634--1644.

\bibitem[{Szegedy et~al.(2015)Szegedy, Liu, Jia, Sermanet, Reed, Anguelov,
  Erhan, Vanhoucke, and Rabinovich}]{szegedy.et.al-2015}
Christian Szegedy, Wei Liu, Yangqing Jia, Pierre Sermanet, Scott Reed, Dragomir
  Anguelov, Dumitru Erhan, Vincent Vanhoucke, and Andrew Rabinovich. 2015.
\newblock Going deeper with convolutions.
\newblock In \emph{Proceedings of the IEEE Conference on Computer Vision and
  Pattern Recognition}, pages 1--9.

\bibitem[{Tellex et~al.(2013)Tellex, Thakerll, Deitsl, Simeonovl, Kollar, and
  Royl}]{tellexll2013toward}
Stefanie Tellex, Pratiksha Thakerll, Robin Deitsl, Dimitar Simeonovl, Thomas
  Kollar, and Nicholas Royl. 2013.
\newblock \href {http://roboticsproceedings.org/rss08/p52.pdf} {Toward
  information theoretic human-robot dialog}.
\newblock \emph{Robotics}, page 409.

\bibitem[{Thomason et~al.(2017)Thomason, Padmakumar, Sinapov, Hart, Stone, and
  Mooney}]{thomason2017opportunistic}
Jesse Thomason, Aishwarya Padmakumar, Jivko Sinapov, Justin Hart, Peter Stone,
  and Raymond~J Mooney. 2017.
\newblock Opportunistic active learning for grounding natural language
  descriptions.
\newblock In \emph{Conference on Robot Learning}, pages 67--76.

\bibitem[{Thomason et~al.(2019)Thomason, Padmakumar, Sinapov, Walker, Jiang,
  Yedidsion, Hart, Stone, and Mooney}]{thomason2019improving}
Jesse Thomason, Aishwarya Padmakumar, Jivko Sinapov, Nick Walker, Yuqian Jiang,
  Harel Yedidsion, Justin Hart, Peter Stone, and Raymond~J Mooney. 2019.
\newblock Improving grounded natural language understanding through human-robot
  dialog.
\newblock \emph{arXiv preprint arXiv:1903.00122}.

\bibitem[{Thomason et~al.(2015)Thomason, Zhang, Mooney, and
  Stone}]{thomason2015learning}
Jesse Thomason, Shiqi Zhang, Raymond~J Mooney, and Peter Stone. 2015.
\newblock \href
  {http://www.aaai.org/ocs/index.php/IJCAI/IJCAI15/paper/download/10957/10931}
  {Learning to interpret natural language commands through human-robot dialog.}
\newblock In \emph{IJCAI}, pages 1923--1929.

\bibitem[{Wiedemeyer(2014 -- 2015)}]{iai_kinect2}
Thiemo Wiedemeyer. 2014 -- 2015.
\newblock {IAI Kinect2}.
\newblock \url{https://github.com/code-iai/iai\_kinect2}.
\newblock Accessed February 28, 2019.

\bibitem[{Yu et~al.(2018)Yu, Lin, Shen, Yang, Lu, Bansal, and
  Berg}]{yu2018mattnet}
Licheng Yu, Zhe Lin, Xiaohui Shen, Jimei Yang, Xin Lu, Mohit Bansal, and
  Tamara~L Berg. 2018.
\newblock Mattnet: Modular attention network for referring expression
  comprehension.
\newblock In \emph{Proceedings of the IEEE Conference on Computer Vision and
  Pattern Recognition}, pages 1307--1315.

\end{thebibliography}
\end{document}